\documentclass[conference]{IEEEtran}
\usepackage{amssymb}
\usepackage{cite}
\ifCLASSINFOpdf
  \usepackage[pdftex]{graphicx}
\else
  \usepackage[dvips]{graphicx}
\fi
\usepackage{algorithmic}
\usepackage{array}
\ifCLASSOPTIONcompsoc
  \usepackage[caption=false,font=normalsize,labelfont=sf,textfont=sf]{subfig}
\else
  \usepackage[caption=false,font=footnotesize]{subfig}
\fi
\usepackage{url}
\usepackage{color}
\usepackage{amsmath,graphicx}
\usepackage{bm}
\usepackage{array}
\usepackage[vlined, linesnumbered, ruled]{algorithm2e}
\usepackage{epsfig}
\usepackage{tikz}
\usepackage{algorithm2e}
\usepackage{pgfplots}
\usepgfplotslibrary{fillbetween}
\usepgfplotslibrary{groupplots}
\usetikzlibrary{arrows}

\newcommand{\Arikan}{Ar{\i}kan }
\newcommand{\Leonardon}{L\'{e}onardon}

\newcolumntype{Y}{>{\centering\arraybackslash}X}

\usepackage[]{parnotes}
\makeatletter
\def\parnoteclear{%
    \gdef\PN@text{}%
    \parnotereset
}
\makeatother
\begin{document}

\title{Using Deep Neural Networks to Predict and Improve the Performance of Polar Codes}

\author{\IEEEauthorblockN{
Mathieu \Leonardon~and
Vincent Gripon
}

\IEEEauthorblockA{IMT Atlantique, Lab-STICC - UMR CNRS 6285}

}

\maketitle

\begin{abstract}
Polar codes can theoretically achieve very competitive Frame Error Rates. In practice, their performance may depend on the chosen decoding procedure, as well as other parameters of the communication system they are deployed upon. As a consequence, designing efficient polar codes for a specific context can quickly become challenging. In this paper, we introduce a methodology that consists in training deep neural networks to predict the frame error rate of polar codes based on their frozen bit construction sequence. We introduce an algorithm based on Projected Gradient Descent that leverages the gradient of the neural network function to generate promising frozen bit sequences. We showcase on generated datasets the ability of the proposed methodology to produce codes more efficient than those used to train the neural networks, even when the latter are selected among the most efficient ones.
\end{abstract}

\section{Introduction}
\label{sec:intro}

Polar Codes are a family of Error Correcting Codes that are used in the 5G NR standard. In order to construct these codes, traditional methods consist in estimating the bit error probabilities with mathematical models. Usually these approaches consider the Successive Cancellation (SC) decoding algorithm as it allows such mathematical formulations.

In practice the Successive Cancellation List (SCL) decoder allows to achieve much better error rates. However SCL is not as easy to manipulate mathematically as SC. Finding good parameters is therefore challenging, and many works have been published with that aim~\cite{Elkelesh2019a,huang2019ai,liao2020construction}.

Most of these methods require estimating bit/frame error rates (BER/FER) using Monte Carlo simulations, and training models to lower these error rates. Ideally, a well trained machine learning algorithm could lead to finding error rates lower than those encountered on its training data.

In this paper, we aim at using neural networks to predict the FER of a polar code from its construction parameters. Once such a neural network is trained, it is possible to use it to generate competitive codes. To this end  we propose to use an adaptation of a classical algorithm typically meant to generate adversarial inputs, i.e. inputs that are specifically designed to fool the neural network decision. In our case, these frozen bit sequences inputs will be generated with the aim at lowering the error rate.

We propose two datasets made of bit frozen sequences and associated FERs, obtained using  Gaussian Approximation or Density Evolution. For each one, we show that it is possible to train neural networks able to predict the FER with high confidence, resulting in an error inflation of about 5\% in average on previously unseen inputs. Using these neural networks, we propose new polar codes achieving lower FERs than the ones used during training. We release both the datasets and the code used in this paper at the following address: \url{https://github.com/brain-bzh/PolarCodesDNN}.


\section{Constructing Polar Codes}
\label{sec:soa}
\subsection{Conventions}
In this paper we consider $(N,K)$ polar codes, where $N$ is the codeblock length and $K$ the
length of the information sequence. We are more precisely interested in polar codes with \Arikan
kernels, as defined in \cite{Arikan2009}. Apart from both $N$ and $K$, such a code
is defined thanks to the frozen bit sequence $\bm{f} \in \{0,1\}^N$. Namely, the 1s in
$\bm{f}=\{f_i\}_{0\leq i < N}$ correspond to the indices of the frozen
bits.

\subsection{Mathematical models}
\label{subsec:models}
Constructing polar codes refers to the way to generate a frozen bits sequence $\bm{f}$,
given $N$, $K$, and a channel model. In \cite{Arikan2009}, construction was
only defined for Binary Erasure Channels. Methods to construct polar codes for
Additive White Gaussian Noise (AWGN) channels were later given in
\cite{Tal2011,Trifonov2012}. All three of this methods sort the
positions of frozen bits, from the most to the least reliable. However, freezing the least reliable bits is not guaranteed to offer the best performance, expressed as the Frame Error Rate (FER) for a targeted SNR. As a matter of fact, finding the best frozen bit sequence (i.e. that yielding the lower Frame Error Rate (FER)), for a given channel, target SNR and decoder, is a difficult challenge that has to be addressed specifically, using complex models or empirical approaches.

\subsection{Learning Approaches}
\label{subsec:approaches}
Some learning-based methods have been recently proposed to further improve the error correcting performance of polar codes. They are based on the mathematical constructions mentioned above and particularize this construction to the specific channel and decoding conditions. In \cite{Elkelesh2019a}, a genetic algorithm is used to modify the frozen bits set in order to improve the performance of polar codes with different channel models (AWGN, Rayleigh) and different decoding algorithms (BP, SCL). Reinforcement Learning (RL) has also been used to address the same problematics \cite{liao2020construction}. The polar code construction is formulated as a maze-traversing game, which is solved using RL methods. This allows to improve the performance of polar codes constructed with \cite{Tal2011} in some of the studied cases.

Deep Learning Techniques have also been used to improve existing codes \cite{huang2019ai}. It is mentioned that neural networks are used to predict performance of a polar code under certain channel and decoding conditions, and some results are provided. However, the methods used to construct and train the network are not described. In this paper, we propose to further investigate the capabilities of neural networks to model and construct polar codes. We give hints about how to construct these neural networks, with detailed experimentations and results. Finally, we propose methods that allow to use neural networks to improve the construction of polar codes for a given channel and decoding algorithm.



\section{Polar Codes Constructing Neural Networks}
\label{sec:method}
\subsection{Dataset Generation}
\label{subsec:dataset}
We propose to train a neural network to predict the FER performance of a polar code. The neural network is trained on a dataset consisting of pairs of frozen
bits sets $\bm{f}$ associated with their corresponding FERs. The FERs are obtained by Monte
Carlo simulations, using the AFF3CT toolbox as a library \cite{Cassagne2019a}.
The frozen bits set is the only parameter that change in the communication chain
accross the simulations. All other parameters are constant, e.g. $N$, $K$,
$E_b/N_0$.

The frozen bits sets space is a high dimensional space $\binom{N}{K}$ in which most
of its elements yield bad FERs. Directly training on poorly performing codes would inevitably lead the neural network to focus predicting high FERs.

In order to generate a relevant subset,
we first use the Gaussian Approximation (GA) method to generate a list
of the frozen bits positions $\bm{p} \in \{0, 1, \dots N-1\}^{N-1}$, that are sorted
according to the reliability that GA associates with each of them. Usually, this sorted
positions are directly used to generate the frozen bits set $\bm{f}$, where
\begin{equation}
f_{p_i} = \left\{
                                          \begin{array}{ll}
                                            1 & \text{if~} i < K \\
                                            0 & \text{if~} i \geq K
                                          \end{array}
                                        \right., \forall 0\leq i < N
\end{equation}
To generate our dataset, a subset of the values in $\bm{p}$ are randomly
shuffled to get a new vector of frozen bits positions $\bm{p^d}$:
\begin{equation}
  p^d_i = \pi^d(p_i), \forall K-r \leq i < K+r
\end{equation}
where $\{\pi^d\}_{0\leq d <D}$ is a set of uniformly distributed random
permutations. The shuffling range $r$ is
chosen empirically. According to our experiments, the best results are obtained
when $r$ is chosen so that there is a one to ten ratio between the worst and the
best measured FER in the generated dataset. In order to get a dataset of size
$D$, the shuffling operation is performed $D$ times, to get $D$ variants of the
frozen bits positions $\bm{p^d}$ and the corresponding frozen bits sets
$\bm{f^d}$.

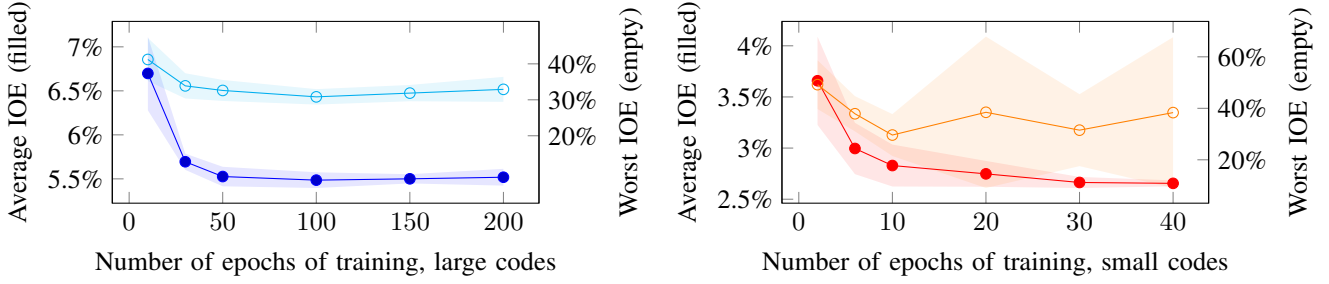
\begin{figure*}[h]
  \begin{center}
    \begin{tikzpicture}
      \begin{axis}[
          width=.4\linewidth, 
          height = 4cm,
          xlabel={Number of epochs of training, large codes}, 
          ylabel={\begin{tabular}{c}Average IOE (filled)\end{tabular}},
          ytick pos=left,
          ytick = {5.5, 6, 6.5, 7},
          yticklabels = {5.5\%, 6\%, 6.5\%, 7\%}
        ]
        \addplot table[x = epoch, y expr = {100*(\thisrow{avg} - 1)}, col sep=comma] {figures/epochs_large.csv};
        \addplot[name path=avg_plus, draw = none] table[x = epoch, y expr={100*((\thisrow{avg} + \thisrow{conf_avg}*1.6134)-1)}, col sep=comma] {figures/epochs_large.csv};
        \addplot[name path=avg_minus, draw = none] table[x = epoch, y expr={100*((\thisrow{avg} - \thisrow{conf_avg}*1.6134)-1)}, col sep=comma] {figures/epochs_large.csv};
        \addplot[blue, opacity = 0.1] fill between [of=avg_minus and avg_plus];
    \end{axis}
        \begin{axis}[
            ylabel={\begin{tabular}{c}Worst IOE (empty)\end{tabular}},
            yticklabel pos=right,
            ytick pos=right,
            ylabel near ticks,
            axis x line=none,
            ymin = 1.25,
            width=.4\linewidth,
            height = 4cm,
            ytick = {20, 30, 40},
            yticklabels = {20\%, 30\%, 40\%}
        ]
        \addplot+[mark=o,draw=cyan, cyan] table[x = epoch, y expr = {100*(\thisrow{worst} - 1)}, col sep=comma] {figures/epochs_large.csv};
        \addplot[name path=worst_plus, draw = none] table[x = epoch, y expr={100*((\thisrow{worst} + \thisrow{conf_worst}*1.6134)-1)}, col sep=comma] {figures/epochs_large.csv};
        \addplot[name path=worst_minus, draw = none] table[x = epoch, y expr={100*((\thisrow{worst} - \thisrow{conf_worst}*1.6134)-1)}, col sep=comma] {figures/epochs_large.csv};
        \addplot[cyan, opacity = 0.1] fill between [of=worst_minus and worst_plus];
      \end{axis}
    \end{tikzpicture}
    \begin{tikzpicture}
      \begin{axis}[
          width=.4\linewidth, 
          height = 4cm,
          xlabel={Number of epochs of training, small codes}, 
          ylabel={\begin{tabular}{c}Average IOE (filled)\end{tabular}},
          ytick pos=left,
          ytick = {2.5, 3, 3.5, 4},
          yticklabels = {2.5\%, 3\%, 3.5\%, 4\%}
        ]
        \addplot+[mark options={fill= red}, red] table[x = epoch, y  expr = {100*(\thisrow{avg} - 1)}, col sep=comma] {figures/epochs_small.csv};
        \addplot[name path=avg_plus, draw = none] table[x = epoch, y expr={100*((\thisrow{avg} + \thisrow{conf_avg}*1.6134)-1)}, col sep=comma] {figures/epochs_small.csv};
        \addplot[name path=avg_minus, draw = none] table[x = epoch, y expr={100*((\thisrow{avg} - \thisrow{conf_avg}*1.6134)-1)}, col sep=comma] {figures/epochs_small.csv};
        \addplot[red, opacity = 0.1] fill between [of=avg_minus and avg_plus];
    \end{axis}
        \begin{axis}[
            ylabel={\begin{tabular}{c}Worst IOE (empty)\end{tabular}},
            yticklabel pos=right,
            ytick pos=right,
            ylabel near ticks,
            axis x line=none,
            width=.4\linewidth,
            height = 4cm,
            ytick = {20, 40, 60},
            yticklabels = {20\%, 40\%, 60\%}
        ]
        \addplot+[mark=o, orange] table[x = epoch, y expr = {100*(\thisrow{worst} - 1)}, col sep=comma] {figures/epochs_small.csv};
        \addplot[name path=worst_plus, draw = none] table[x = epoch, y expr={100*((\thisrow{worst} + \thisrow{conf_worst}*1.6134)-1)}, col sep=comma] {figures/epochs_small.csv};
        \addplot[name path=worst_minus, draw = none] table[x = epoch, y expr={100*((\thisrow{worst} - \thisrow{conf_worst}*1.6134)-1)}, col sep=comma] {figures/epochs_small.csv};
        \addplot[orange, opacity = 0.1] fill between [of=worst_minus and worst_plus];
      \end{axis}
     \end{tikzpicture}
    \caption{Evolution of validation Inflation Of Errors depending on the number of epochs used for training, in both the case of large codes (left) and small codes (right). Standard deviation obtained on 10 runs is also shown.}
  \end{center}
  \label{epochs}
\end{figure*}

\subsection{Neural network}
\label{subsec:nn}

Let us first recall that a neural network can be modelled as a mathematical function that is obtained by assembling elementary subfunctions called \emph{layers}. A typical layer is a function of the form $F^\ell: \bm{x}\mapsto \sigma(\bm{W}^\ell\bm{x} + \bm{b}^\ell)$, where $\bm{W}^\ell$ and $\bm{b^\ell}$, respectively a matrix and a vector, are made of values that are trainable parameters, and $\sigma$ is a nonlinear predetermined function which acts on each component of its input independently. The number of dimensions in the input $\bm{x}$ is denoted $d_\ell^{input}$ and the number of dimensions in the output $F_\ell(\bm{x})$ is denoted $d_\ell^{output}$. Assembling layers can be performed by performing composition, additions, concatenations...

In our work, we consider very simple architectures that are defined using three hyperparameters:
\begin{itemize}
    \item The \emph{number of layers} $L$. We denote each associated layer function $(F_\ell)_{1\leq\ell\leq L}$. We use Rectified Linear Units (ReLU) as nonlinear functions in all layer functions, except the last for which we do not use a nonlinear function.
    \item The \emph{number of neurons in hidden layers}, denoted $H$. We then fix the dimensions: $d_1^{input}=N$, $(d_\ell^{input} = H, \forall \ell > 1)$, $(d_\ell^{output} = H, \forall \ell < L)$, $d_L^{output} = 1$.
    \item The \emph{shortcut gap} $G$ defined in analogy to celebrated ResNet architectures~\cite{he2016deep} (see details below).
\end{itemize}

Once all parameters $L, H$ and $G$ are fixed, we can define the neural network function $F$. Denote $\bm{f}$ an input sequence of bits. We compute the following sequence:
\newcommand{\Mod}[1]{\ (\mathrm{mod}\ #1)}
\begin{equation}
    \left\{\begin{array}{rl} \bm{f}^0 &= \bm{f}\\\bm{f}^{\ell+1} &= \left\{\begin{array}{ll}F_{\ell+1}(\bm{f}^{\ell}) + \bm{f}^{\ell+1 - G}&\text{if }\exists k \in \mathbb{N}^*,\ell = k G\\&\text{and } \ell+1 < L\\F_{\ell+1}(\bm{f}^{\ell})&\text{otherwise}\end{array}\right.\end{array}\right.
\end{equation}

Then, we define $F(\bm{f}) = \bm{f}^L, \forall{\bm{f}}$.

\begin{figure}[h]
  \begin{center}
    \begin{tikzpicture}[thick]
      \node(i) at (0,0) {$\bm{f}$};
      \node(o) at (8,0) {$F(\bm{f})$};
      \node[inner sep = 0pt](plus) at (5,0) {$\bigoplus$};
      \tikzstyle{every node} = [draw, rectangle, minimum height = 2cm, minimum width = 0.8cm]
      \node(fc1) at (1,0) {\tiny{$F_1$}};
      \node(fc2) at (2.3,0) {\tiny{$F_2$}};
      \node(fc3) at (4,0) {\tiny{$F_{G+1}$}};
      \tikzstyle{every node} = []
      \small{
        \path[->,>=stealth']
        (i) edge (fc1)
        (fc1) edge (fc2)
        (fc2) edge[dashed] (fc3)
        (fc3) edge (plus)
        (plus) edge[dashed] (o)
        (4.9, 1.5) edge (4.9, 0.2);
      }
      \path[->,>=stealth',dashed]
      (5.1, 1.5) edge (6,1.5);
      \draw[]
      (1.65, 0) -- (1.65, 1.5) -- (4.9, 1.5);
      \draw[dashed]
      (5.1, 0.2) -- (5.1, 1.5);
    \end{tikzpicture}
  \end{center}

  \caption{Depiction of the architectures used to predict the FER based on input frozen bit sequences. The direct flow of data from input to output is periodically combined with shortcuts of length $G$.}
  \label{architecture}
\end{figure}
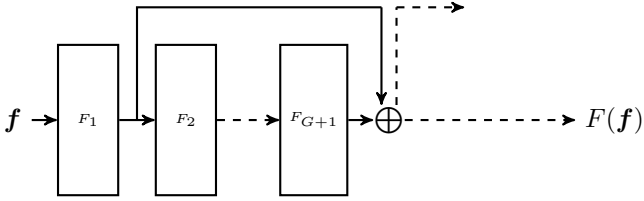

In particular, when $G$ is no lesser than $L-1$, this sequence boils down to composing $F = F^L \circ F^{L-1} \circ \dots \circ F^1$. A toy depiction of this architecture is shown in Figure~\ref{architecture}.

To train $F$ to associate an input bit sequence $\bm{f}$ with the corresponding FER denoted $\text{FER}_{\bm{f}}$, we use a variant of the Stochastic Gradient Descent algorithm named Adam~\cite{kingma2014adam}, which is meant to minimize the Mean Square Error between $F(\bm{f})$ and $\text{FER}_{\bm{f}}$ in logscale. In other words, we aim at approximately solving:
\begin{equation}
    \arg\min_{\{\bm{W}^\ell, \bm{b}^\ell\}_{1\leq \ell\leq L}}\mathbb{E}[(F(\bm{f}) - \log(\text{FER}_{\bm{f}}))^2].
    \label{loss}
\end{equation}

We use a logscale to better encompass for the range of FERs we expect to observe in applications.

\subsection{Inflation Of Error (IOE)}

To train a neural network, we make use of a training set $T$ and a validation set $V$, both made of pairs $(\bm{f}, \text{FER}_{\bm{f}})$. The idea is to define these sets so that $V \cap T = \emptyset$. Once a neural network has been trained using $T$ as a proxy to the expectation of Equation~\eqref{loss}, we evaluate on $V$ its ability to generalize predictions to previously unseen inputs.

We define the Inflation Of Error (IOE) of $(\bm{f}, \text{FER}_{\bm{f}})\in V$ for the neural network function $F$ as:
\begin{equation}
    \text{IOE}(\bm{f}, F) = \max\left\{\frac{\text{FER}_{\bm{f}}}{\exp(F(\bm{f}))}, \frac{\exp(F(\bm{f}))}{\text{FER}_{\bm{f}}}\right\} - 1.
\end{equation}

As an example, an IOE of 100\% means that the FER predicted using the neural network is half or twice the actual FER of the considered validation sample.

In our experiments, we are typically interested in measuring the average IOE or the worst (i.e. maximal) IOE on the validation set.

\subsection{Constructing codes using neural networks}

Once neural networks have been trained to predict FER based on input frozen bit sequences, we can leverage them to propose new efficient codes.

We adapt the methodology described in~\cite{madry2017towards}, called Projected Gradient Descent (PGD). PGD has been widely used in the context of adversarial attacks, where the aim is to generate inputs meant to fool the neural network predictions. The idea is to fix the neural network parameters, and use gradient descent to update inputs so that it translates the corresponding output towards new decisions. In our case, the goal is to use the prediction of the neural network as the function to minimize, since it means lowering the FER.

In more details, we implement a straight-through~\cite{bengio2013estimating} procedure, in which we manipulate inputs that are not necessarily binary: such inputs can be seen as a relaxation of bit sequences into real-valued ones. When we estimate the FER using the neural network function, we manipulate a quantized version of the input, where all values below the median are set to one bit and all others the other one. We compute the gradient with respect to this input, but apply it to its real-valued version. More details are available in Algorithm~\ref{pgd}. This algorithm comes with two parameters: a number of iterations $I$ and a gradient step $\mu$.

\begin{algorithm}
$\bm{f} \leftarrow$ random binary initialization\\
\For{\text{a fixed number of iterations} $I$}{
$\bm{\tilde{f}} \leftarrow \text{quantized version of } \bm{f}$\\
$y \leftarrow F(\bm{\tilde{f}})$\\
\For{$0\leq i < N$}{
$f_i \leftarrow f_i - \mu \frac{\partial y}{\partial \tilde{f}_i}$
}
}
\text{Return quantized version of} $\bm{f}$\\
\caption{Algorithm used to generate low FER polar codes.}
\label{pgd}
\end{algorithm}

\section{Experiments}
\label{sec:results}

Throughout this section, we consider two types of polar codes, namely $(256,128)$, termed \emph{small} and $(1024,512)$, termed \emph{large}.  In both cases, an SCL decoder is used, with a list depth of respectively $\mathcal{L}=4$ and $\mathcal{L}=32$. The SNR values used for any construction method are respectively $E_b/N_0=3.2dB$ and $E_b/N_0=2.7dB$. For the small codes, we generated a total of $77'466$ frozen bit sequences $\bf{f}$, split arbitrarily in 80\% used for training and 20\% for validation. For the large codes, we generated a total of $15'862$ frozen bit sequences, with the same proportions for training and validation.


We use classical data standardization techniques, where we center and reduce both input and output components. On the input, this has the effect of replacing 0s with -1s. We also remove coordinates in the inputs that are constant on our datasets. We end up using 36 out of the 256 input dimensions for small codes and 112 out of the 1024 input dimensions for large codes. The minimal FER in the training and validation sets are 1.67e-4 for small codes and 5.75e-5 for large ones. As a reference for upcoming experiments, we also computed the average IOE and worst IOE obtained when using a constant predictor (predicting the average of outputs). We obtained: average IOE: 86.03\% and worst IOE: 1541.32\% for small codes and average IOE: 47.90\% and worst IOE: 4716.20\% for large ones. These quantities can be thought of as ``chance levels''. For all reported results, we compute at least an average over 10 runs, where each run is obtained with different randomly drawn initial weights for the layer parameters, and different order of presenting data during training.

\subsection{Effect of hyperparameters on IOE}

As a first series of experiments, we wanted to empirically study the impact of hyperparameters in our architectures on the average and worst IOEs. We first randomly searched for an efficient starting set of hyperparameters, and then looked at the influency of varying a single parameter, keeping the other constant. We ended up using 20 epochs for training for small codes and 100 epochs for training for large ones. In both cases, we used 320 neurons in the hidden layers, a depth of 6 and a shortcut gap of 3.

\textbf{Prediction performance of neural networks:}

On small codes, we obtain after 10 runs the following performance: average IOE: 2.65\% $\pm$ 0.06\%, worst IOE: 29.36\% $\pm$ 5.63\% for small codes (confidence intervals obtained at 95\%). For the large codes, we obtain: average IOE: 5.46\% $\pm$ 0.04\%, worst IOE: 31.28\% $\pm$ 0.04\%. In other words, the ratio between predicted FER and actual one is no more than 6\% in average for our datasets, which is way smaller than the chance level we estimated. Also, in the worst case, the actual FER is about 30\% larger or smaller than the actual one, which demonstrates the outstanding generalization ability of the trained neural networks. It is also worth pointing out that the estimation of FER using Monte Carlo simulations is subject to a small error, and that this error could be significant with respect to the average IOE measured in our experiments.

\textbf{Effect of the number of epochs in training:}

In Figure~\ref{epochs}, we report the average and worst IOEs obtained depending on the number of epochs used to train the neural network architectures. As expected the small codes require less epochs to reach convergence, as the dataset is way larger than in the case of large codes. Interestingly, we observe a small increase of IOEs when training large codes for too many epochs, which can indicate a small risk of overfitting.

\textbf{Effect of the number of neurons in hidden layers:}

\begin{figure}[h]
  \begin{center}
    \begin{tikzpicture}
      \begin{axis}[
          width=.85\linewidth, 
          height = 4cm,
          xlabel={Size of hidden layers}, 
          ylabel={\begin{tabular}{c}Average IOE (filled)\end{tabular}},
          ytick pos=left,
          yticklabels={{2\%}, {4\%}, {6\%}, {8\%}, {10\%}},
          ytick = {2, 4, 6 ,8, 10}
        ]
        \addplot table[x = size, y expr={100 * (\thisrow{avgl} - 1)}, col sep=comma] {figures/sizes.csv};
        \addplot[name path=avg_plusl, draw = none] table[x = size, y expr={100 * ((\thisrow{avgl} + \thisrow{conf_avgl}*1.6134) - 1)}, col sep=comma] {figures/sizes.csv};
        \addplot[name path=avg_minusl, draw = none] table[x = size, y expr={100 * ((\thisrow{avgl} - \thisrow{conf_avgl}*1.6134) - 1)}, col sep=comma] {figures/sizes.csv};
        \addplot[blue, opacity = 0.1] fill between [of=avg_minusl and avg_plusl];
        \addplot+[mark=*,mark options={fill= red}, red] table[x = size, y expr ={100 * (\thisrow{avgs} - 1)}, col sep=comma] {figures/sizes.csv};
        \addplot[name path=avg_pluss, draw = none] table[x = size, y expr={100 * ((\thisrow{avgs} + \thisrow{conf_avgs}*1.6134)-1)}, col sep=comma] {figures/sizes.csv};
        \addplot[name path=avg_minuss, draw = none] table[x = size, y expr={100 * ((\thisrow{avgs} - \thisrow{conf_avgs}*1.6134)-1)}, col sep=comma] {figures/sizes.csv};
        \addplot[red, opacity = 0.1] fill between [of=avg_minuss and avg_pluss];
    \end{axis}
        \begin{axis}[
            ylabel={\begin{tabular}{c}Worst IOE (empty)\end{tabular}},
            yticklabel pos=right,
            ytick pos=right,
            ylabel near ticks,
            axis x line=none,
            ymin=1.,
            ytick={20, 40, 60, 80, 100},
            yticklabels={20\%, 40\%, 60\%, 80\%, 100\%},
            width=.85\linewidth,
            height = 4cm,
        ]
        \addplot+[mark=o,draw=cyan, cyan] table[x = size, y expr = {100*(\thisrow{worstl} - 1)}, col sep=comma] {figures/sizes.csv};
        \addplot[name path=worst_plusl, draw = none] table[x = size, y expr={100 * ((\thisrow{worstl} + \thisrow{conf_worstl}*1.6134) - 1)}, col sep=comma] {figures/sizes.csv};
        \addplot[name path=worst_minusl, draw = none] table[x = size, y expr={100 * ((\thisrow{worstl} - \thisrow{conf_worstl}*1.6134) - 1)}, col sep=comma] {figures/sizes.csv};
        \addplot[cyan, opacity = 0.1] fill between [of=worst_minusl and worst_plusl];
        \addplot+[mark=o, orange, ] table[x = size, y expr = {100 * (\thisrow{worsts} - 1)}, col sep=comma] {figures/sizes.csv};
        \addplot[name path=worst_pluss, draw = none] table[x = size, y expr={100 * ((\thisrow{worsts} + \thisrow{conf_worsts}*1.6134) - 1)}, col sep=comma] {figures/sizes.csv};
        \addplot[name path=worst_minuss, draw = none] table[x = size, y expr={100 * ((\thisrow{worsts} - \thisrow{conf_worsts}*1.6134) - 1)}, col sep=comma] {figures/sizes.csv};
        \addplot[orange, opacity = 0.1] fill between [of=worst_minuss and worst_pluss];
      \end{axis}
    \end{tikzpicture}
    \caption{Evolution of validation Inflation Of Errors depending on the size of the hidden layers in the trained architectures, for large codes (blue and cyan) and small codes (red and orange). Standard deviation obtained on 10 runs is also shown.}
  \end{center}
  \label{size}
\end{figure}
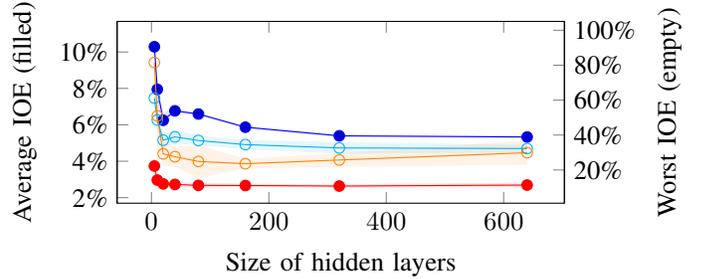

Next, we wanted to study the impact of the number of neurons in hidden layers $H$ in the architecture. In Figure~\ref{size}, we depict the evolution of IOEs as a function of $H$. Except for the small outlier obtained with the blue curves -- which is probably due to the limited number of runs --, we observe an expected global decrease of IOE as the number of neurons grow. A plateau is soon reached around $H = 300$. Interestingly, the worst IOE for small codes seems to increase with the largest values of $H$, again making one suspect about potential overfitting.

\textbf{Effect of the depth of the architecture:}

\begin{figure}[h]
  \begin{center}
    \begin{tikzpicture}
      \begin{axis}[
          width=.85\linewidth, 
          height = 4cm,
          xlabel={Depth of trained architectures}, 
          ylabel={\begin{tabular}{c}Average IOE (filled)\end{tabular}},
          ytick pos=left,
          ytick = {3, 4, 5, 6},
          yticklabels = {3\%, 4\%, 5\%, 6\%},
        ]
        \addplot table[x = depth, y expr = {100 * (\thisrow{avgl} - 1)}, col sep=comma] {figures/depth.csv};
        \addplot[name path=avg_plusl, draw = none] table[x = depth, y expr={100 * ((\thisrow{avgl} + \thisrow{conf_avgl}*1.6134) - 1)}, col sep=comma] {figures/depth.csv};
        \addplot[name path=avg_minusl, draw = none] table[x = depth, y expr={100 * ((\thisrow{avgl} - \thisrow{conf_avgl}*1.6134) - 1)}, col sep=comma] {figures/depth.csv};
        \addplot[blue, opacity = 0.1] fill between [of=avg_minusl and avg_plusl];
        \addplot+[mark = *, mark options={fill= red}, red] table[x = depth, y expr= {100 * (\thisrow{avgs}-1)}, col sep=comma] {figures/depth.csv};
        \addplot[name path=avg_pluss, draw = none] table[x = depth, y expr={100 * ((\thisrow{avgs} + \thisrow{conf_avgs}*1.6134) - 1)}, col sep=comma] {figures/depth.csv};
        \addplot[name path=avg_minuss, draw = none] table[x = depth, y expr={100 * ((\thisrow{avgs} - \thisrow{conf_avgs}*1.6134) - 1)}, col sep=comma] {figures/depth.csv};
        \addplot[red, opacity = 0.1] fill between [of=avg_minuss and avg_pluss];
    \end{axis}
        \begin{axis}[
            ylabel={\begin{tabular}{c}Worst IOE (empty)\end{tabular}},
            yticklabel pos=right,
            ytick pos=right,
            ylabel near ticks,
            axis x line=none,
            width=.85\linewidth,
            height = 4cm,
            ytick = {20, 40, 60, 80},
            yticklabels = {20\%, 40\%, 60\%, 80\%}
        ]
        \addplot+[mark=o,draw=cyan, cyan] table[x = depth, y expr = {100 * (\thisrow{worstl} - 1)}, col sep=comma] {figures/depth.csv};
        \addplot[name path=worst_plusl, draw = none] table[x = depth, y expr={100 * ((\thisrow{worstl} + \thisrow{conf_worstl}*1.6134) - 1)}, col sep=comma] {figures/depth.csv};
        \addplot[name path=worst_minusl, draw = none] table[x = depth, y expr={100 * ((\thisrow{worstl} - \thisrow{conf_worstl}*1.6134)-1)}, col sep=comma] {figures/depth.csv};
        \addplot[cyan, opacity = 0.1] fill between [of=worst_minusl and worst_plusl];
        \addplot+[mark=o, orange] table[x = depth, y expr = {100*(\thisrow{worsts} - 1)}, col sep=comma] {figures/depth.csv};
        \addplot[name path=worst_pluss, draw = none] table[x = depth, y expr={100 * ((\thisrow{worsts} + \thisrow{conf_worsts}*1.6134) - 1)}, col sep=comma] {figures/depth.csv};
        \addplot[name path=worst_minuss, draw = none] table[x = depth, y expr={100 * ((\thisrow{worsts} - \thisrow{conf_worsts}*1.6134) - 1)}, col sep=comma] {figures/depth.csv};
        \addplot[orange, opacity = 0.1] fill between [of=worst_minuss and worst_pluss];
      \end{axis}
    \end{tikzpicture}
  \label{fig:depth}
    \caption{Evolution of validation Inflation Of Errors depending on the depth of the trained architectures, for large codes (blue and cyan) and small codes (red and orange). Standard deviation obtained on 10 runs is also shown.}
  \end{center}
\end{figure}
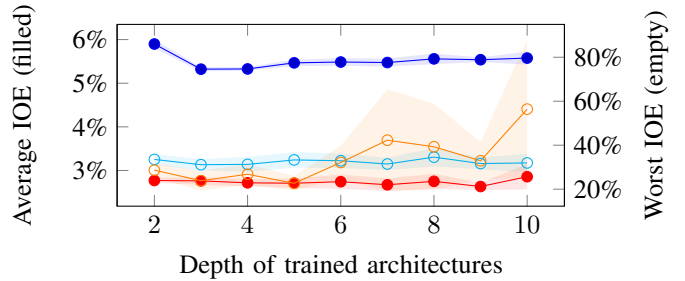

In Figure~\ref{fig:depth}, we vary the depth of trained architectures and report the corresponding IOEs. The effect of the depth seems to play a role analogous to that of the number of neurons in hidden layers, where too small or too large values can cost higher IOEs. The best obtained average IOEs are for depth 3 and 9 respectively, with a pretty insignificant effect between these values. On the contrary, increasing the depth seems to cause way larger worst IOEs for small codes.

\textbf{Effect of the shortcut gap length:}

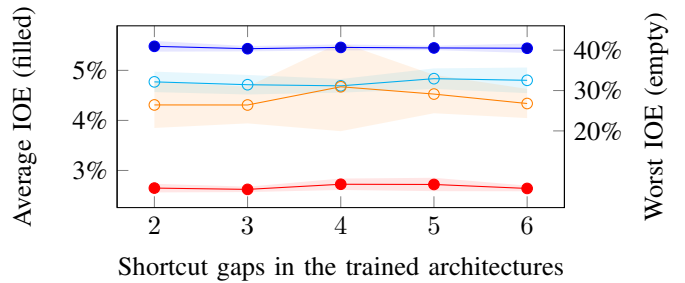
\begin{figure}[]
  \begin{center}
    \begin{tikzpicture}
      \begin{axis}[
          width=.85\linewidth, 
          height = 4cm,
          xlabel={Shortcut gaps in the trained architectures}, 
          ylabel={\begin{tabular}{c}Average IOE (filled)\end{tabular}},
          ytick pos=left,
          ytick = {3, 4, 5},
          yticklabels ={3\%, 4\%, 5\%}
        ]
        \addplot table[x = gap, y expr = {100 * (\thisrow{avgl} - 1)}, col sep=comma] {figures/gaps.csv};
        \addplot[name path=avg_plusl, draw = none] table[x = gap, y expr={100 * ((\thisrow{avgl} + \thisrow{conf_avgl}*1.6134) - 1)}, col sep=comma] {figures/gaps.csv};
        \addplot[name path=avg_minusl, draw = none] table[x = gap, y expr={100 * ((\thisrow{avgl} - \thisrow{conf_avgl}*1.6134) - 1)}, col sep=comma] {figures/gaps.csv};
        \addplot[blue, opacity = 0.1] fill between [of=avg_minusl and avg_plusl];
        \addplot+[mark=*,mark options={fill= red}, red] table[x = gap, y expr = {100*(\thisrow{avgs} - 1)}, col sep=comma] {figures/gaps.csv};
        \addplot[name path=avg_pluss, draw = none] table[x = gap, y expr={100 * ((\thisrow{avgs} + \thisrow{conf_avgs}*1.6134)-1)}, col sep=comma] {figures/gaps.csv};
        \addplot[name path=avg_minuss, draw = none] table[x = gap, y expr={100*((\thisrow{avgs} - \thisrow{conf_avgs}*1.6134)-1)}, col sep=comma] {figures/gaps.csv};
        \addplot[red, opacity = 0.1] fill between [of=avg_minuss and avg_pluss];
    \end{axis}
        \begin{axis}[
            ylabel={\begin{tabular}{c}Worst IOE (empty)\end{tabular}},
            yticklabel pos=right,
            ytick pos=right,
            ylabel near ticks,
            axis x line=none,
            ymin=1.,
            width=.85\linewidth,
            height = 4cm,
            ytick = {20, 30, 40},
            yticklabels = {20\%, 30\%, 40\%}
        ]
        \addplot+[mark=o,draw=cyan, cyan] table[x = gap, y expr = {100*(\thisrow{worstl} - 1)}, col sep=comma] {figures/gaps.csv};
        \addplot[name path=worst_plusl, draw = none] table[x = gap, y expr={100*((\thisrow{worstl} + \thisrow{conf_worstl}*1.6134)-1)}, col sep=comma] {figures/gaps.csv};
        \addplot[name path=worst_minusl, draw = none] table[x = gap, y expr={100*((\thisrow{worstl} - \thisrow{conf_worstl}*1.6134)-1)}, col sep=comma] {figures/gaps.csv};
        \addplot[cyan, opacity = 0.1] fill between [of=worst_minusl and worst_plusl];
        \addplot+[mark=o, orange] table[x = gap, y expr = {100*(\thisrow{worsts} - 1)}, col sep=comma] {figures/gaps.csv};
        \addplot[name path=worst_pluss, draw = none] table[x = gap, y expr={100 * ((\thisrow{worsts} + \thisrow{conf_worsts}*1.6134)-1)}, col sep=comma] {figures/gaps.csv};
        \addplot[name path=worst_minuss, draw = none] table[x = gap, y expr={100 * ((\thisrow{worsts} - \thisrow{conf_worsts}*1.6134)-1)}, col sep=comma] {figures/gaps.csv};
        \addplot[orange, opacity = 0.1] fill between [of=worst_minuss and worst_pluss];
      \end{axis}
    \end{tikzpicture}
    \caption{Evolution of validation Inflation Of Errors depending on the shortcut gaps in the trained architectures, for large codes (blue and cyan) and small codes (red and orange). Standard deviation obtained on 10 runs is also shown.}
  \end{center}
  \label{gaps}
\end{figure}

Next, we tested the effect of the shortcut gap $G$ on the IOE of trained architectures. Obtained results are presented in Figure~\ref{gaps}. Of all tested parameters, the shortcut gap length seems to be the one having the smallest influence on obtained IOEs. It is worth noting that a depth of 6 is equivalent to not using shortcuts in our architectures. When zooming, we still observe a significant improvement when using small gaps in both the case of small and large codes.

\textbf{Using standard methods to increase predicted IOEs:}

Finally, we implemented standard techniques from the field of Deep Learning meant to either prevent overfitting or improve prediction performance. We implemented DropOut~\cite{srivastava2014dropout}, including on the input~\cite{devries2017improved}, where we randomly mask coordinates of input or output vectors computed throughout the architecture during training, BatchNorms~\cite{ioffe2015batch}, which center and reduce each dimension of computed vectors during training, and learn corresponding coefficients to be used on validation data, and Mixup~\cite{zhang2017mixup}, a simple data-augmentation procedure in which training inputs are linearly interpolated and trained to associate the corresponding linear interpolation of the outputs. Results are presented in Table~\ref{techniques}. Interestingly, none of the tested methods resulted in improved IOEs on our tests. We suspect that this is because these techniques are mostly meant to be applied when dealing with raw input signals, such as in the case of vision or audio signals. In our case, the discrete and combinatorial nature of inputs might make these techniques unsuitable.


\def\checkmark{\tikz\fill[scale=0.4](0,.35) -- (.25,0) -- (1,.7) -- (.25,.15) -- cycle;}
\begin{table}
    \caption{Influence of common techniques in Deep Learning on the average IOE (confidence intervals at 95\% are indicated): Dropout, Batch-Norms (BNs) and Mixup.}
    \centering
    \begin{tabular}{|c|c|c|c|c|}
    \hline
         DropOut & BNs & Mixup & IOE (Large) & IOE (Small)\\
        \hline
        \hline
         & & & \textbf{5.46\%} ($\pm$ 0.04\%) & \textbf{2.65\%} ($\pm$ 0.06\%)\\
         \hline
         & & \checkmark & 7.69\% ($\pm$ 0.06\%)& 3.08\% ($\pm$ 0.17\%)\\
         \hline
         & \checkmark & & 8.95\% ($\pm$ 0.37\%) & 6.42\% ($\pm$ 0.80\%)\\
         \hline
         & \checkmark & \checkmark & 9.14\% ($\pm$ 0.50\%)& 8.05\% ($\pm$ 0.11\%)\\
         \hline
         \checkmark & & & 36.78\% ($\pm$ 0.28\%) & 58.92\% ($\pm$ 0.37\%)\\
         \hline
         \checkmark & & \checkmark & 37.43\% ($\pm$ 0.44\%) & 60.78\% ($\pm$ 0.69\%)\\
         \hline
         \checkmark & \checkmark & & 32.77\% ($\pm$ 0.23\%) & 52.96\% ($\pm$ 0.21\%)\\
         \hline
         \checkmark & \checkmark & \checkmark & 32.89\% ($\pm$ 0.33\%) & 53.31\% ($\pm$ 0.18\%)\\
         \hline
    \end{tabular}
    \label{techniques}
\end{table}

\subsection{Constructing Codes}
\label{subsec:constr_results}

After having explored the impact of hyperparameters on the IOEs, we fixed parameters for constructing interesting codes. We chose: $(L,H,G) = (3, 640, 3)$ trained for 100 epochs for large codes (i.e. not using shortcuts) and $(L,H,G) = (5, 320, 3)$ trained for 40 epochs for small codes. When using Algorithm~\ref{pgd}, we used a maximum of 5000 iterations and a gradient step $\mu = 0.1$.

\begin{figure}[t]
  \begin{tikzpicture}
    \begin{groupplot}[/pgfplots/table/ignore chars={|}, 
                width=0.5\linewidth, height=0.400\linewidth,
                group style={group name=fer, group size= 2 by 1, horizontal sep=2cm, vertical sep=2.2cm},
                xlabel=$E_b/N_0$ (dB), grid=both, grid style={gray!30},
                ylabel=Frame Error Rate,
                xmin=0.8, xmax=4.2,
                ymin=0.000002,
                ymode = log,
                tick align=outside, tickpos=left, 
                legend pos=south west, legend columns=1]
        \nextgroupplot[ylabel=FER,xmin=0.8, xmax=6.0]
        \addplot[name path=small_ga,    mark=o,      blue,  semithick] table [x=Eb/N0, y=FER] {figures/fer/small_ga.csv};    \label{plot:line5}
        \addplot[name path=small_worst, mark=none,   blue,  semithick] table [x=Eb/N0, y=FER] {figures/fer/small_worst.csv}; \label{plot:line6}
        \addplot[name path=small_best,  mark=none,   blue,  semithick] table [x=Eb/N0, y=FER] {figures/fer/small_best.csv};  \label{plot:line7}
        \addplot[name path=small_nn,    mark=o,      red,   semithick] table [x=Eb/N0, y=FER] {figures/fer/small_nn.csv};    \label{plot:line8}
        \addplot[blue,  opacity = 0.1] fill between [of=small_worst and small_best];
        \nextgroupplot[ylabel=, xshift=-0.6cm]
        \addplot[name path=large_ga,    mark=o,      blue, semithick] table [x=Eb/N0, y=FER] {figures/fer/large_ga.csv};    \label{plot:line1}
        \addplot[name path=large_worst, mark=none,   blue, semithick] table [x=Eb/N0, y=FER] {figures/fer/large_worst.csv}; \label{plot:line2}
        \addplot[name path=large_best,  mark=none,   blue, semithick] table [x=Eb/N0, y=FER] {figures/fer/large_best.csv};  \label{plot:line3}
        \addplot[name path=large_nn,    mark=o,      red,  semithick] table [x=Eb/N0, y=FER] {figures/fer/large_nn.csv};    \label{plot:line4}
        \addplot[blue, opacity = 0.1] fill between [of=large_worst and large_best];

        \coordinate (legend) at (axis description cs:0.755,0.03);
    \end{groupplot}

  \end{tikzpicture}
  \caption{Frame Error Rate performance of small (left) and large (right) codes, with the GA construction (blue circles), the generated datasets (blue areas), and our proposed construction (red circles).}
  \label{fer}
\end{figure}
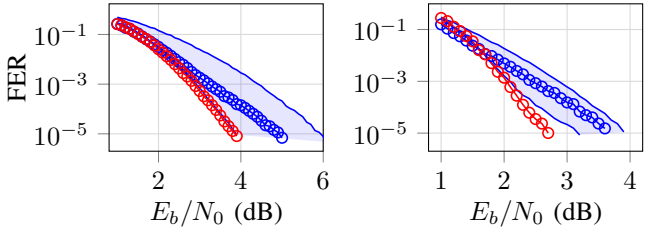

With large codes, we obtained a frozen bit sequence reaching as low as 1.01e-5 as FER, validated using Monte Carlo simulations, which is way lower than the minimum observed on the training set (5.75e-5). With small codes, we obtained several interesting candidates that reached competitive FERs, but not surpassing by a large margin the examples of the training set. This was expected due to the relative small size of these codes and the abundance of already very optimized bit sequences in the training set. The corresponding FERs of best candidates are depicted in Figure~\ref{fer}.

\section{Conclusion}
\label{sec:conclu}
In this paper, we investigated how to build neural networks capable of predicting the error correction performance of polar codes. We proposed an algorithm to extract competitive codes from trained neural networks.
Two datasets of frozen bits sets associated with their corresponding FER values, for a fixed SNR, were generated for two different code lengths, under SCL decoding. These datasets were used to train neural networks to predict the FER of any frozen bits set. We obtained in average a ratio of errors of less than 1.06 for the (1024,512) codes and 1.02 for (256,128) codes. The codes we generated using trained neural networks were shown to perform better than those of the training datasets. Finally, the source code used in this article as well as the generated datasets are published for reproducibility and reuse.

\bibliographystyle{IEEEtran}
\bibliography{nn_polar}

\end{document}